\documentclass{Interspeech2024}
\usepackage{multirow}
\usepackage{booktabs}
\usepackage{amssymb}
\usepackage{pifont}
\usepackage{caption}
\usepackage{colortbl}  
\usepackage{xcolor}

\interspeechcameraready

\title{Cross-Modal Denoising: A Novel Training Paradigm for Enhancing Speech-Image Retrieval
\thanks{*Corresponding author.}}

\name{Lifeng}{Zhou}
\name[affiliation={*}]{Yuke}{Li}
\name{Rui}{Deng}
\name{Yuting}{Yang}
\name{Haoqi}{Zhu}

\address{
  Netease Yidun  AI Lab, Hangzhou, China}
\email{\{hzzhoulifeng,liyuke,dengrui01,yangyuting04,zhuhaoqi\}@corp.netease.com}


\begin{document}

\maketitle

\begin{abstract}
    
The success of speech-image retrieval relies on establishing an effective alignment between speech and image. 
Existing methods often model cross-modal interaction through simple cosine similarity of the global feature of each modality, which fall short in capturing fine-grained details within modalities.
To address this issue, we introduce an effective framework and
a novel learning task named cross-modal denoising (CMD) to
enhance cross-modal interaction to achieve finer-level cross-modal alignment. Specifically, CMD is a denoising task designed to reconstruct semantic features from noisy features within one modality by interacting features from another modality.
Notably, CMD operates exclusively during model training and can be removed during inference without adding extra inference time.
The experimental results demonstrate that our framework outperforms the state-of-the-art method by 2.0\% in mean R@1 on the Flickr8k dataset and by 1.7\% in mean R@1 on the SpokenCOCO dataset for the speech-image retrieval tasks, respectively.
These experimental results validate the efficiency and effectiveness of our framework.  
\end{abstract}

\section{Introduction}
By harnessing plentiful labeled data and computational resources, speech processing systems have demonstrated remarkable performance \cite{Synnaeve_Xu}\cite{Wang_Mohamed_Le_Liu_Xiao_Mahadeokar}.
Unfortunately, the scarcity of labeled data for the majority of languages, coupled with the expensive nature of transcribing large volumes of speech data, has fueled a rising interest in the development of techniques capable of extracting valuable insights from unlabeled data \cite{Badino_Canevari_Fadiga_Metta_2014}\cite{Bhati_Nayak_Murty_2017}.

Recently, there has been a notable emergence of self-supervised learning (SSL) methods as a prominent strategy for acquiring representations from unlabeled audio data, as evidenced by studies such as \cite{Liu_Yang_Chi_Hsu_Lee_2020}, \cite{Liu_Chung_Glass_2020}, and \cite{Baevski_Zhou_Mohamed_Auli_2020}. These methods have garnered attention for their effectiveness in this area, as demonstrated by the work of \cite{Synnaeve_Xu} and \cite{Schneider_Baevski_Collobert_Auli_2019}.
Furthermore, the exploration of multimodal data and the extraction of valuable information from it have been investigated as an alternative approach to improving the performance of speech processing systems.
Pairing images with speech has been widely utilized to improve speech processing, ultimately resulting in the advancement of visually grounded speech (VGS) models, as exemplified in the work of \cite{Peng_Harwath}. 
These models have demonstrated their utility across a range of applications, such as speech recognition \cite{Hsu_Harwath_Glass_2019}, \cite{Peng_Huang_Gu_Xie_Wang_Jiao_Ye_2021}, and \cite{He_Sainath}, word discovery \cite{Harwath_Glass_2015}, and multilingual spoken language processing \cite{harwath2018vision}. Typically, VGS models undergo training and evaluation in the context of speech-image retrieval tasks.

The advancement of VGS models has led to a substantial improvement in the accuracy of speech-image retrieval systems, underscoring the significant potential of speech-image retrieval as a standalone application.
In the FaST-VGS method \cite{9747103fast-vgs}, the authors utilize an innovative training and retrieval approach that integrates dual-encoder and cross-attention architectures, enabling a single model to achieve both rapid and accurate speech-image retrieval capabilities. 
SpeechCLIP, as detailed in \cite{shih2022speechclip}, uses a speech encoder initialized with a pre-trained SSL model \cite{Chen_Wang_Chen_Wu_Liu_Chen_Li_Kanda_Yoshioka_Xiao_et} and aligns it with a frozen CLIP image encoder using paired speech-image data. This enables SpeechCLIP to achieve SOTA performance in speech-image retrieval tasks.

While effective, these methods have limitations. For instance, in FaST-VGS, using an object detector as the image encoder can limit expressive power due to the constraints of the detector and its predefined visual vocabulary.
SpeechCLIP replaces the object detector with an image encoder from CLIP to extract image features. It encodes speech and images separately, utilizing contrastive learning as the training objective. However, in contrastive learning, the interaction between modalities is managed solely through the cosine similarity of the speech and image features, which may pose challenges in achieving fine-grained alignment. As a result, this approach may lead to false positive matching during inference when images and speech share similar semantics but differ in details.

\begin{figure*}[ht]
  \centering
  \includegraphics[width=\linewidth]{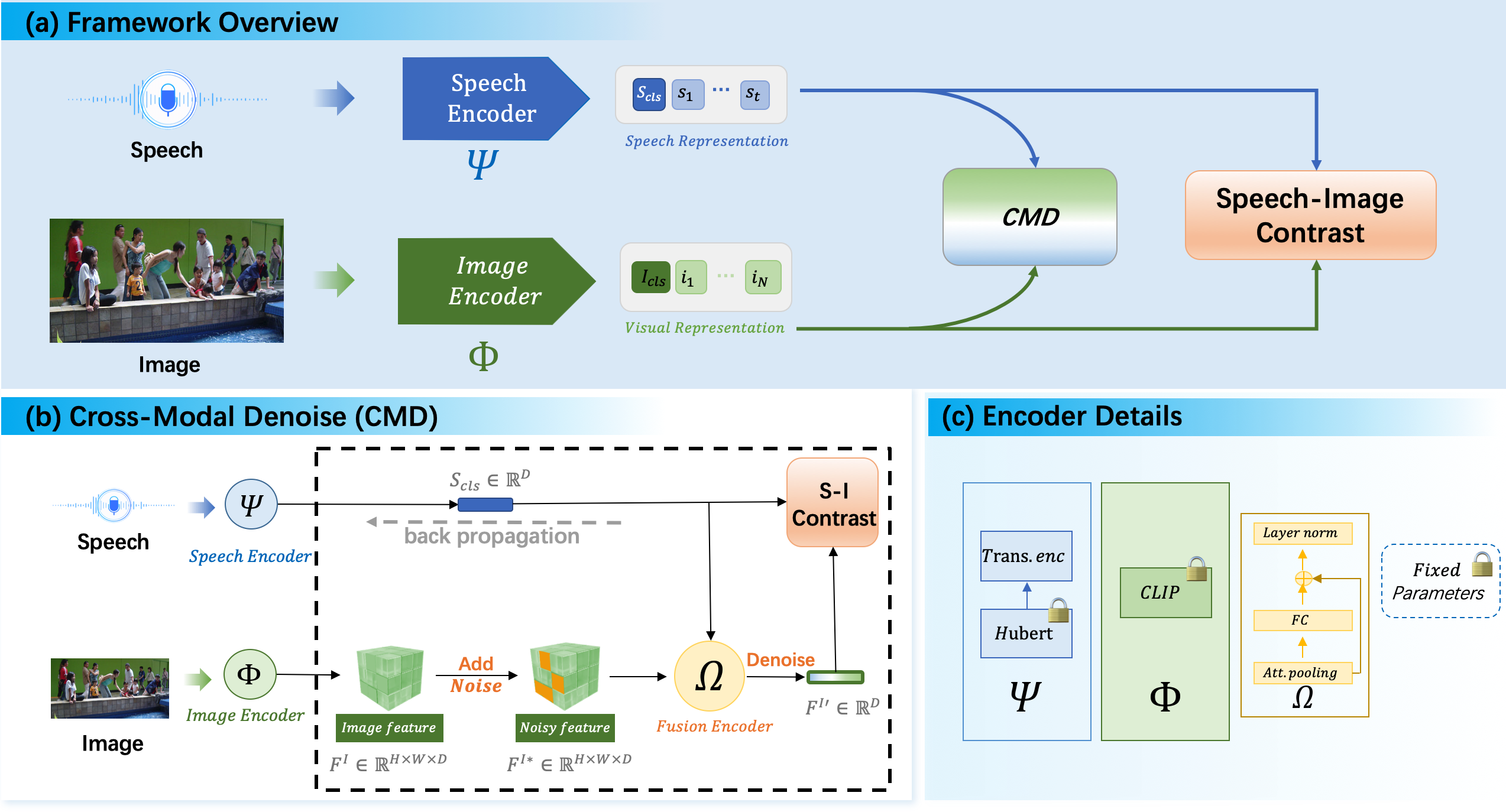}
  \caption{The overview of our proposed framework. Figure (a) showcases that our framework is optimized with speech-image contrastive learning tasks and CMD tasks. Figure (b) provides details of the CMD tasks, while Figure (c) presents the specifics of the three encoders used in our framework. 
 }
  \label{fig:architecture}
  \vspace{-2.0em}
\end{figure*}

Therefore, designing a better interaction between modalities to facilitate fine-grained alignment between modalities is of crucial significance for the performance of speech-image retrieval systems.
This paper introduces an innovative framework and a novel learning task cross-modal denoising (CMD) to enhance cross-modal interaction and achieve fine-grained cross-modal alignment. CMD is a denoising task designed to reconstruct semantic features from noisy features within one modality by interacting features from another modality. 
The objective of the CMD is to enhance speech representations, enabling them to focus on specific image-patch contexts, thereby achieving fine-grained cross-modal alignment.

Our main contributions can be summarized as follows:
\begin{itemize}
\item  We propose a simple yet powerful framework with only 14M trainable parameters to achieve effective alignment between speech and image, ultimately leading to more accurate speech-image retrieval.

\item We introduce a novel cross-modal learning task CMD to enhance cross-modal fusion, thereby achieving fine-grained cross-modal alignment. Importantly, CMD operates exclusively during model training and can be removed during inference without adding extra inference time.

\item Our framework has exhibited a significant improvement of 2.0\% in mean R@1 on the benchmark dataset of Flickr8k Audio Capitions Coupus and 1.7\% in mean R@1 on the SpokenCOCO dataset, surpassing the performance of the current state-of-the-art approach. 
\vspace{-1.0em}
\end{itemize}

.

\section{Methods}
\subsection{Preliminaries}
In this section, we will present a concise overview of the two pre-trained models used in our framework: HuBERT and CLIP. 

\noindent\textbf{Hidden-unit BERT (HuBERT)} \cite{hsu2021hubert}
is a self-supervised learning speech model that employs a masked prediction objective, akin to the well-known BERT \cite{Devlin_Chang_Lee_Toutanova_2019} model. By predicting masked speech frames based on the surrounding context, it captures essential speech representations. The model architecture includes a CNN feature extractor coupled with a transformer encoder, enabling it to effectively extract meaningful speech features for various downstream tasks \cite{Yang_Chi}.

\noindent\textbf{CLIP} \cite{radford2021learning} leverages contrastive learning to pre-train visual models at a large scale using natural language supervision \cite{Savarese}\cite{blip}\cite{Li_Selvaraju_Gotmare_Joty_Xiong_Hoi_2021}, which is derived from paired image-text data. By employing two separate encoders for processing images and text, CLIP seeks to align semantically similar images and text captions. This enables CLIP to seamlessly transfer across a range of computer vision tasks with minimal supervision.

In our framework, pre-trained HuBERT and CLIP models are frozen and serve as feature extractors.

\subsection{Architecture} 
As illustrated in Figure \ref{fig:architecture}, we employ the speech encoder $\Psi$ and the image encoder $\Phi$ to extract speech features $S=\{S_{cls}, s_1, ..., s_T\}$ and image features $I=\{I_{cls}, i_1, ..., i_N\}$, where $S_{cls}$ and $I_{cls}$ represent the normalized global speech and image semantic features, and $s_{i}$ and $i_{i}$ represent the frame-level speech feature and patch-level image feature.
The framework is optimized with multitasks, including speech-image contrastive learning and CMD tasks. The speech-image contrastive learning task is designed to align the speech and image features at a coarse level. The CMD task aims to achieve fine-grained alignment between speech and image modality. 

As shown (b) in Figure \ref{fig:architecture}, CMD can be viewed as a combination of feature denoising and speech-image contrastive learning tasks. Specifically,  we first introduce noise to the image features $F^{I}$ to get noisy image features $F^{I*}$, where $F^{I}\in  \mathbb{R}^{H\times W\times D}$ is reshaped from patch-level image features ${i_1, ..., i_N}$. The noise addition process is as follows: for a given image feature $F^{I}$, we randomly select certain patch features and then replace them with patch features from other image features within the same batch.
Subsequently, the noisy image features $F^{I*}$ along with the speech semantic features $S_{cls}$ interact in the multimodal fusion encoder $\Omega$ to reconstruct image global semantic feature $F^{I'}$. 
This process can be presented as follows: 
\begin{equation}
Attn(I^{*}\mid s)=Softmax\left(\frac{Q_s K_{I^{*}}^T}{\sqrt{D}}\right) V_{I^{*}},
\end{equation}
\begin{equation}
F^{I'}=LN\left(FC\left(Attn\left(I^{*} \mid s\right)\right)+Attn\left(I^{*} \mid s\right)\right)^T,
\end{equation}
where $Q, K, V$ denote the query, key, and value embeddings,
$Softmax$ refers to the normalization function, $LN$ stands for the layer normalization layer, and $FC$ represents a fully connected layer. 
The objective of the multimodal fusion encoder is to improve speech representations, enabling them to concentrate on specific image-patch contexts in order to achieve fine-grained cross-modal alignment.
Lastly, we use speech-image contrastive learning to put paired speech semantic feature $S_{cls}$ and denoised image semantic feature $F^{I'}$ close together in the latent
space and to pull them apart from the other features. 

\begin{table*}[ht]
    \centering
    \setlength{\tabcolsep}{3.42mm}{
    \begin{tabular}{c c c c c c}
      \toprule
      Speech Encoder & Speech Encoder & Image Encoder  &  Fusion Encoder & Trainable Params &  Total Params\\
      \hline
      Hubert Large & Transformer encoder & VIT-L/14 & FC & \textbf{\multirow{2}{*}{14M} }&   \multirow{2}{*}{752M} \\
      (316M) & \textbf{(13.4M)} & (422M) & \textbf{(0.6M)} &  &  \\
    \bottomrule
    \end{tabular}}
    \caption{The model details of our framework.} 
       \label{tab:details} 
    \vspace{-2.5em}
\end{table*}

The details of the three encoders used in our framework are shown (c) in Figure \ref{fig:architecture}, the speech encoder $\Psi$ comprise a self-supervised learning speech model Hubert \cite{hsu2021hubert} and 1 layer transformer encoder.  Drawing inspiration from SUPERB \cite{Yang_Chi}, we integrate the CNN output of HuBERT with the hidden representations from its transformer encoder using learnable weights. This weighted combination forms a sequence of speech features. These features, along with the CLS token, are subsequently fed into the transformer encoder to extract speech embeddings, denoted as $S=\{S_{cls}, s_1, ..., s_N\}$.
Additionally, we utilize the image encoder of CLIP \cite{radford2021learning} as $\Phi$ to extract image features.  
The multimodal fusion encoder $\Omega$ comprises an attentive pooling layer, a fully connected layer, and a layer normalization layer. Notably, the parameters of Hubert in $\Psi$ and $\Phi$ are fixed during training process.

\subsection{Training Objectives}
Our model is trained with two primary objectives: speech-image contrastive learning on the unimodal encoders and CMD on the multimodal fusion encoder.

\noindent\textbf{Speech-Image Contrastive Learning} aims to align speech and image features at a coarse level, facilitating cross-modal learning in the multimodal fusion encoder. It involves learning a similarity function $s = S_{cls}^T I_{cls}$, ensuring that parallel speech-image pairs receive higher similarity scores than non-parallel pairs.

\begin{table*}[ht]
    \centering
    \setlength{\tabcolsep}{3.42mm}{
    \begin{tabular}{l r r r r r r r r r}
      \toprule
      \multirow{ 2}{*}{Method} & \multicolumn{3}{c}{ Speech $\rightarrow$ Image } & \multicolumn{3}{c}{ Image $\rightarrow$ Speech } & \multicolumn{3}{c}{ Mean }\\
      \cline{2-4} \cline{5-7} \cline{8-10}
      & R@1 & R@5 & R@10 & R@1 & R@5 & R@10 & R@1 & R@5 & R@10\\
      \hline  & \multicolumn{9}{c}{ Flickr8k } \\
\hline $\rm {FaST\text{-}VGS_{CO}}$ \cite{9747103fast-vgs} & 26.6 & 56.4 & 68.8 & 36.2 & 66.1 & 76.5  & 31.4 & 61.3 & 72.6 \\
\hline $\rm {FaST\text{-}VGS_{CTF}}$ \cite{9747103fast-vgs} & 29.3 & 58.6 & 71.0 & 37.9 & 68.5 & 79.9  & 33.6 & 63.6 & 75.5 \\
\hline MILAN \cite{Sanabria_Waters_Baldridge_2021}& 33.2 & 62.7 & 73.9 & 49.6 & 79.2 & 87.5  & 41.4 & 71.0 & 80.7 \\
\hline Cascaded SpeechCLIP \cite{shih2022speechclip}& 14.7 & 41.2 & 55.1 & 21.8 & 52.0 & 67.7 & 18.3 & 46.6 & 61.4 \\ 
\hline Parallel SpeechCLIP \cite{shih2022speechclip}& 39.1 & 72.0 & 83.0 & 54.5 & 84.5 & 93.2 & 46.8 & 78.3 & 88.1\\
\rowcolor[HTML]{F0F0F0} 
\hline Ours & \textbf{40.7} & \textbf{75.1} & \textbf{85.8} & \textbf{56.8} & \textbf{86.2} & \textbf{94.2} & \textbf{48.8} & \textbf{80.7} & \textbf{90.0} \\ 
\hline & \multicolumn{9}{c}{ SpokenCOCO } \\
\hline ResDAVEne \cite{Hsu_Harwath_Glass_2019} & 17.3 & 41.9 & 55.0 & 22.0 & 50.6 & 65.2 & 19.65 & 46.3 & 60.1\\
\hline $\rm {FaST\text{-}VGS_{CO}}$ \cite{9747103fast-vgs}  & 31.8 & 62.5 & 75.0 & 42.5 & 73.7 & 84.9 & 37.2 & 68.1 & 80.0\\
\hline $\rm {FaST\text{-}VGS_{CTF}}$ \cite{9747103fast-vgs} & 35.9 & 66.3 & 77.9 & 48.8 & 78.2 & 87.0 & 42.4 & 72.3 & 82.5\\
\hline Cascaded SpeechCLIP \cite{shih2022speechclip}& 6.4 & 20.7 & 31.0 & 9.6 & 27.7 & 39.7 & 8.0 & 24.2 & 35.4\\
\hline Seg. SpeechCLIP \cite{Bhati_Villalba_Moro-Velazquez_Thebaud_Dehak_2023}& 28.2 & 55.3 & 67.5 & 28.5 & 56.1 & 68.9 & 28.4 & 55.7 & 68.2\\
\hline Parallel SpeechCLIP \cite{shih2022speechclip}& 35.8 & 66.5 & 78.0 & 50.6 & 80.9 & 89.1 & 43.2 & 73.7 & 83.5 \\
\rowcolor[HTML]{F0F0F0}
\hline Ours & \textbf{37.5} & \textbf{67.3} & \textbf{78.6} & \textbf{52.3} & \textbf{81.4} & \textbf{89.7} & \textbf{44.9} & \textbf{74.4} & \textbf{84.2}\\
    \bottomrule
    \end{tabular}}
    \caption{Recall scores for speech-image retrieval on Flickr8k and SpokenCOCO testing sets.} 
       \label{tab:main-recall} 
    \vspace{-1.5em}
\end{table*}

\begin{table*}[ht]
    \centering
    \begin{tabular}{l r r r r r r r r r}
      \toprule
      \multirow{2}{*}{Method} & \multicolumn{3}{c}{ Speech $\rightarrow$ Image } & \multicolumn{3}{c}{ Image $\rightarrow$ Speech } & \multicolumn{3}{c}{ Mean }\\
      \cline{2-4} \cline{5-7} \cline{8-10}
      & R@1 & R@5 & R@10 & R@1 & R@5 & R@10 & R@1 & R@5 & R@10\\
\hline supervised  & 40.7 & 75.1 & 85.8 & 56.8 & 86.2 & \textbf{94.2} & 48.8 & 80.7 & 90.0 \\
\hline zero-shot & \textbf{48.9} & \textbf{78.3} & \textbf{87.5} & \textbf{61.4} & \textbf{88.2} & 93.8  & \textbf{55.1} & \textbf{83.2} & \textbf{90.7} \\ 
    \bottomrule
    \end{tabular}  
    \caption{Recall scores for zero-shot speech-image retrieval on Flickr8k testing sets.} 
    \label{tab:zs-recall} 
    \vspace{-2.0em}
\end{table*}

For each speech, we calculate the softmax-normalized similarity between the speech and image features as follows:
\begin{equation}
p_j^{\mathrm{s} 2 \mathrm{i}}(S)=\frac{\exp \left(s\left(S, I_j\right) / \tau\right)}{\sum_{j=1}^B \exp \left(s\left(S, I_j\right) / \tau\right)},
\label{ps2i}
\end{equation}
where $\tau$ is a learnable temperature parameter, $B$ is the mini-batch size.
For each image, the softmax-normalized image and speech similarity is calculated as:
\begin{equation}
p_j^{\mathrm{i} 2 \mathrm{s}}(I)=\frac{\exp \left(s\left(I, S_j\right) / \tau\right)}{\sum_{j=1}^B \exp \left(s\left(I, S_j\right) / \tau\right)}.
\end{equation}
Let $\boldsymbol{y}^{\mathrm{s} 2 \mathrm{i}}(S)$ and $\boldsymbol{y}^{\mathrm{i} 2 \mathrm{s}}(I)$ denote the ground-truth one-hot similarity vectors, where negative pairs are assigned a probability of 0 and positive pairs are assigned a probability of 1. The speech-image contrastive loss is then defined as the cross-entropy $\mathrm{H}$ between the predicted probabilities $\boldsymbol{p}$ and the ground-truth similarities $\boldsymbol{y}$, as follows:

\begin{equation}
\mathcal{L}_{\mathrm{sic}}=\frac{1}{2}[\mathrm{H}\left(\boldsymbol{y}^{\mathrm{s} 2 \mathrm{i}}(S), \boldsymbol{p}^{\mathrm{s} 2 \mathrm{i}}(S)\right)\\+\mathrm{H}\left(\boldsymbol{y}^{\mathrm{i} 2 \mathrm{s}}(I), \boldsymbol{p}^{\mathrm{i} 2 \mathrm{s}}(I)\right)]
\label{eq:cl}
\end{equation}

\noindent\textbf{Cross-modal denoising} is a denoising task which aims to align the speech and image features at a fine-grained level.
As illustrated in Figure \ref{fig:architecture}(b), CMD can be viewed as a
combination of feature denoising and speech-image contrastive
learning tasks. 
The training loss for CMD is similar with Equation \ref{eq:cl} as follows:
\begin{equation}
\mathcal{L}_{\mathrm{cmd}}=\frac{1}{2}[\mathrm{H}\left(\boldsymbol{y}^{\mathrm{s} 2 \mathrm{i}}(S), \boldsymbol{p}^{\mathrm{s} 2 \mathrm{i'}}(S)\right)+\mathrm{H}\left(\boldsymbol{y}^{\mathrm{i} 2 \mathrm{s}}(I), \boldsymbol{p}^{\mathrm{i'} 2 \mathrm{s}}(I)\right)].
\label{eq:cl1}
\end{equation}
The full pre-training objective of our framework is donated as:
\begin{equation}
\mathcal{L}=\mathcal{L}_{\mathrm{sic}}+\alpha\mathcal{L}_{\mathrm{cmd}},
\end{equation} 
where $\alpha$ is a hyper-parameter used to balance $\mathcal{L}_{\mathrm{sic}}$ and $\mathcal{L}_{\mathrm{cmd}}$.

\section{Experiment}

\subsection{Setup} 
\noindent\textbf{Dataset}. 
Our model is trained and evaluated on speech-image retrieval tasks using the Flickr8k Audio Captions Corpus \cite{Harwath_Glass_2015} and the SpokenCOCO dataset \cite{Hsu_Harwath_Miller_Song_Glass_2021}. In both datasets, each image is paired with five spoken captions, which are human utterances of text captions. The Flickr8k dataset comprises 8k images and 46 hours of speech, while SpokenCOCO includes 123k images and 742 hours of speech. Consistent with FaST-VGS \cite{9747103fast-vgs}, we utilize the Karpathy \cite{Karpathy_Fei-Fei_2017} split for the SpokenCOCO dataset.

\noindent\textbf{Setup}.
The speech encoder $\Psi$ consists of Hubert and a single-layer transformer encoder. The Hubert model utilized in our experiments is Hubert-Large, while the transformer encoder has eight attention heads, and the hidden dimension of the transformer encoder is the same as that of HuBERT. As for the CLIP image encoder $\Phi$, we used ViT-L/14. Both the parameters of HuBERT and CLIP are kept frozen throughout the training process. Additionally, the input and output dimensions of the fully connected layer utilized in the fusion encoder $\Omega$ are both set at 768. For detailed model configurations, please refer to Table \ref{tab:details}.
During the noise addition process, we randomly select 30\% of image patch-level features to add noise.
Given that both datasets have 5 speech captions for each image, we modify the ground-truth labels for contrastive learning to account for multiple positives during training. Each positive sample is assigned a ground-truth probability of $1/5$. 
All models are trained using the Adam optimizer with a weight decay of $10^{-6}$, a batch size of 128, and a total of 60k training steps. The learning rate is linearly increased to $10^{-4}$ during the first 4k steps and then gradually decreased to $10^{-8}$. All experiments are performed on a machine equipped with 8 32GB V100 GPUs
During inference, we remove the multimodal fusion encoder $\Omega$ and only compute the feature similarity score between speech and image semantic features $S_{cls}$ and $I_{cls}$ for all speech-image pairs.  

\noindent\textbf{Evaluation Metric}.
To evaluate the cross-modal retrieval performance of our framework, we use the widely adopted Recall at K (R@K) metric, where higher values indicate better performance. We report the results for both speech-to-image retrieval and image-to-speech retrieval

\subsection{Speech-Image Retrieval}
In this section, we assess the performance of our framework in speech-image retrieval tasks, thereby demonstrating the effectiveness of our models in aligning speech with image features.
The cross-modal retrieval performance of our method is presented in Table \ref{tab:main-recall}. In comparison to previous methods, we have achieved the best retrieval performance in both speech-to-image retrieval and image-to-speech retrieval tests. Our model has shown significant improvements over the previous best model \cite{shih2022speechclip}, with increases of 2.0\% in mean R@1, 2.4\% in mean R@5, and 1.9\% in mean R@10 on the Flickr8k dataset. Besides, our model has demonstrated improvements of 1.7\% in mean R@1, 0.7\% in mean R@5, and 0.7\% in mean R@10 on the SpokenCOCO dataset. These improvements can be primarily attributed to our model's ability, achieved through joint training with contrastive learning and CMD tasks, to identify shared semantics between images and speech while also capturing their subtle differences.

\subsection{Zero-Shot Speech-Image Retrieval}
To assess the generalization ability of our framework, we performed zero-shot retrieval by directly evaluating the model trained on SpokenCOCO with the testing sets of Flickr8K, marking the first exploration of the generalization capability of a speech-image retrieval model to the best of the authors' knowledge.
The results, shown in Table \ref{tab:zs-recall}, indicate that the \emph{supervised} model trained on the Flickr8k training sets is significantly outperformed by the model trained on the SpokenCOCO training sets. This highlights the excellent generalization ability of our model. The superior performance can be attributed to the model being trained on the larger SpokenCOCO dataset compared to the smaller Flickr8k dataset, demonstrating the model's scalability.

\subsection{Ablation Studies}
In this section, we conduct ablation
studies and report the results in mean R@1 on two datasets for simplicity.

\noindent\textbf{Effectiveness of CMD.} 
Table \ref{tab:cmd} studies the effect of CMD on cross-modal retrieval. In comparison to training without CMD, the inclusion of the CMD training task resulted in a 2.3\% improvement on the Flickr8k dataset and a 1.9\% improvement on the SpokenCOCO dataset, indicating the effectiveness of CMD.

\begin{figure}[]
  \setlength{\abovecaptionskip}{-0.1pt}
  \centering
  \begin{minipage}{0.45\textwidth}
    \centering
    \includegraphics[width=\linewidth]{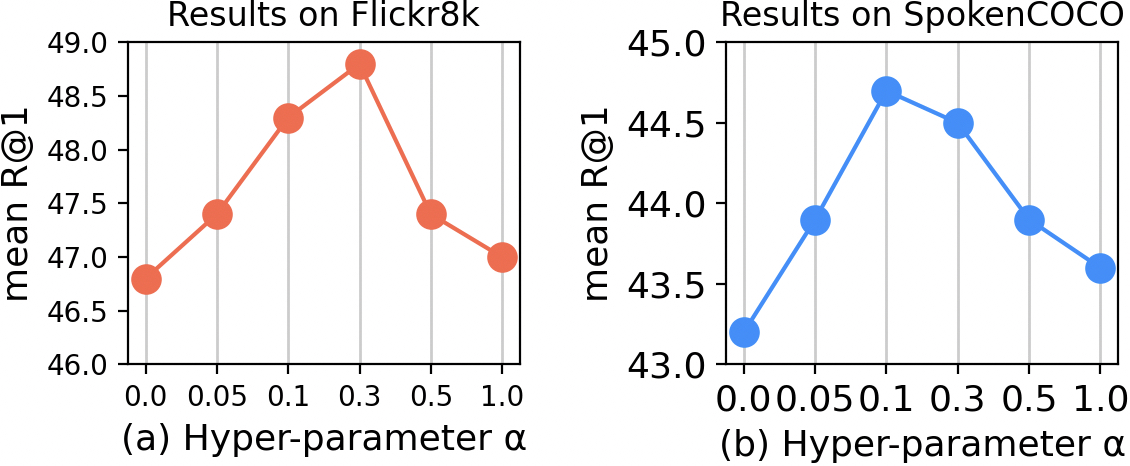}
    \caption{ Effect of the balance hyper-parameter $\alpha$}
    \label{fig:hyper}
  \end{minipage}
  \vspace{-1.0em}
\end{figure}

\noindent\textbf{The balance hyper-parameter $\alpha$.} The hyper-parameter determines the weight of CMD task. 
To assess its impact, we explore various scale ranges for $\alpha$ within the interval of [0.0, 1.0] on two datasets. The results, depicted in Figure \ref{fig:hyper}, indicate that the optimal value of $\alpha$ varies across different datasets.
\begin{table}[t]
    \centering
    \begin{tabular}{lcc}
        \hline Training task & Flickr8k & SpokenCOCO \\
        \hline w/o CMD & 46.5 & 43.0 \\
        \hline w/ CMD & $\textbf{48.8}$ & $\textbf{44.9}$ \\
        \hline
    \end{tabular}
    \caption{Ablation study of CMD} 
    \label{tab:cmd}
    \vspace{-2.0em}
\end{table}

\section{Conclusions}
We propose a simple yet powerful framework to enhance the alignment between speech and image, thereby leading to more accurate speech-image retrieval. The framework is trained with cross-modal contrastive learning and cross-modal denoising (CMD) tasks. Specifically, CMD is a novel denoising task designed to enhance speech representations, enabling them to focus on specific image-patch contexts, thereby achieving fine-grained cross-modal alignment.
Importantly, CMD operates solely during model training and can be removed during inference without adding any inference time. The experimental results demonstrate that our framework outperforms the state-of-the-art method by 2.0\% in mean R@1
on the Flickr Audio Captions Corpus and by 1.7\% in mean R@1 on
the SpokenCOCO dataset for the speech-image retrieval tasks,
respectively. These experimental results validate the efficiency
and effectiveness of our framework. 
In our future works, we aim to continue advancing speech-image retrieval performance, as the accuracy of speech-image retrieval systems has lagged behind their image-text counterparts.

\ifinterspeechfinal
\else
\fi

\bibliographystyle{unsrt}
\bibliography{mybib}

\end{document}